# Tagging multimedia stimuli with ontologies

Marko Horvat[1], Siniša Popović[1], Nikola Bogunović[2] and Krešimir Ćosić[1]

[1]Department of Electric Machines, Drives and Automation
[2]Department of Electronics, Microelectronics, Computer and Intelligent Systems
Faculty of Electrical Engineering and Computing, University of Zagreb
Unska 3, HR-10000 Zagreb, Croatia
Phone: (385) 1- 6129 521  Fax: (385) 1-6129 705  E-mail: marko.horvat2@fer.hr

**Abstract** - Successful management of emotional stimuli is a pivotal issue concerning Affective Computing (AC) and the related research. As a subfield of Artificial Intelligence, AC is concerned not only with the design of computer systems and the accompanying hardware that can recognize, interpret, and process human emotions, but also with the development of systems that can trigger human emotional response in an ordered and controlled manner. This requires the maximum attainable precision and efficiency in the extraction of data from emotionally annotated databases While these databases do use keywords or tags for description of the semantic content, they do not provide either the necessary flexibility or leverage needed to efficiently extract the pertinent emotional content. Therefore, to this extent we propose an introduction of ontologies as a new paradigm for description of emotionally annotated data. The ability to select and sequence data based on their semantic attributes is vital for any study involving metadata, semantics and ontological sorting like the Semantic Web or the Social Semantic Desktop, and the approach described in the paper facilitates reuse in these areas as well.

## I. INTRODUCTION

Tagging is a process of annotation or assignment of metadata to information. While annotation can be simply described as adding a note by way of comment or explanation [1] it actually entails significantly more complexity and the accompanying theory or research. Tag can be strictly defined as a 'free-text keyword. Thus tagging becomes an 'indexing process for assigning tags to resources' [2]. A shared collection of tags used within a particular system is called folksonomy. Folksonomies are also referred to as collaborative tagging, social classification, social indexing, and social tagging.

Development and management of folksonomies, particularly maintenance of their semantics and descriptive values, is an important and demanding task which compromises two distinctive sets of problems: i) problems with local variations and ii) problems with distributed variations. By their very nature tags carry small amounts of information, i.e. semantics, but have many variations which are the result of a large spectrum of issues like spelling, professional and cultural background, but also individual psychological processes [3]. Subsequently, a set of tags does not always have to correctly and consistently represent mental model of its every user. While these problems are related to the local frame of reference, the problems with distributed variations are resulting from the fact that the majority of tagging systems manage and interpret tag semantics in different ways. It is often difficult to find correlations between semantically similar but formally, i.e. lexically, different tags, and thus it becomes a challenge to connect and aggregate metadata from different tagging applications.

All these issues are ubiquitous to the tagging process itself. Therefore, they also apply to the usage of annotated, or tagged, emotional stimuli in the fields of Affective Computing (AC) and Artificial Intelligence (AI). Broadly speaking, our work in AC is concerned with psychotherapy and psychological training. We aim to develop an integrated system for automated adaptation of virtual reality (VR) based scenarios driven by the subject's physiology, with a rationale regarding application of the system in the Posttraumatic stress disorder (PTSD) treatment [4][5]. A part of our efforts is directed at the design and development of the emotional stimuli generator which is able to actuate contextually and temporarily anticipated emotional responses in a human subject [6]. In order to do this the generator is able to display still images and video clips, generate specific sounds, and display synthetic virtual environments based on the control commands from the system's supervisor. In this process it is imperative that the commands result in emotionally and contextually aligned stimuli which are individually conformed to a specific subject and his present mental state. We are using International Affective Picture System (IAPS) and International Affective Digitized Sounds (IADS) databases which store annotated emotional stimuli. The databases use free-text keywords, or tags, to describe the meaning of individual stimuli. However, the keywords are semantically scattered, taxonomically disordered, and subsequently cumbersome to use or extract information from.

In this paper we will explain how to introduce ontology-based tagging in the existing emotionally annotated databases to achieve twofold improvement: a) more efficient extraction of knowledge, and b) higher informative, i.e. semantically descriptive, stimuli data value. We will also demonstrate this concept with StimuliGenerator (StimGen) computer application that we have developed as a part of work on the scientific project "Adaptive Control of Scenarios in VR Therapy of PTSD"[†].

The next chapter introduces emotionally annotated databases, their benefits and drawbacks. The third chapter comments why it is necessary to move away from keywords, or tags, towards ontologies in description of emotionally annotated data. Construction of ontologies for description of emotional stimuli semantics is explained in the fourth chapter. Finally, merging of additional data like psychological model values and author's information with the developed ontology is explained in the fifth chapter.

[†]This research has been partially supported by the Croatian Ministry of Science, Education and Sports.

## II. EMOTIONALLY ANNOTATED DATABASES

Annotated databases have simple yet effective structure. They typically consist of a file system conjoined with a corresponding manifesto that enumerates and describes all files in the database. The manifesto is a simple text, CSV (comma-separated value) or any other easily readable file in which each row describes one singular piece of data. Often only one manifesto carries all metadata, but it is possible to add more manifests and cross-link them if needed. Data rows are guaranteed unique and although primary or foreign keys aren't explicitly declared an annotated database can be transformed into a relation database. In the case of IAPS or IADS the value of the primary key can be the name of the stimuli file which is always an unique four digit number, e.g. 5999.jpg, 6000.jpg, 100.wav, 101.wav, etc. The files have URI (Uniform Resource Identificator) which makes it easy to fetch them and use programmatically. Such design principles and nomenclature make annotated databases technologically rudimentary by today's standards, but for the same reasons also robust and good cross-platform solutions.

Emotional annotated databases, e.g. IAPS or IADS, use storage mechanisms which are generic to all annotated databases, but with the single purpose of aggregating emotional content. IAPS and IADS store emotional content according to the quantitative-dimensional psychological model. This model numerically describes the meaning of the emotional impulse in a 2D plane with the respect to the axis of valence or pleasure (x-axis; denoted $pl$) and the axis of excitation or arousal (y-axis; denoted $ar$). Values of both axis are normalized in interval [1, 9]. Fig. 1. shows IAPS data projected on the referent 2D plane with each dot representing one emotionally annotated picture.

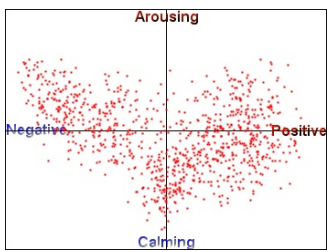

Fig. 1. IAPS data in the referent quantitative-dimensional 2D plane.

The semantic context of the stimuli in such model is described with a closed set of keywords. One stimulus is always tagged with one keyword. In other words, all IAPS and IADS stimuli have to be tagged, but with no more than one keyword. Fig. 2. displays 4 stimuli that are tagged with the keyword *Boat*.

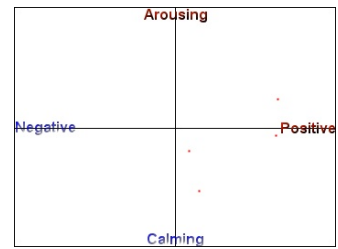

Fig. 2. Selection of IAPS data tagged with keyword *Boat*.

Quantitative-dimensional models correlate better with biological-emotional changes then discrete categorical models which categorize emotions in slots such as joy, sadness, surprise, disgust, fear and anger. For these reasons, quantitative-dimensional models are considered to be suitable for describing motive in psychotherapy and psychological training where is necessary to control the emotional physiology of a subject with stimuli. An example of annotated visual emotional stimuli from IAPS with ($pl$, $ar$) = {(5.59, 2.88), (5.34, 4.23), (7.48, 4.74), (7.53, 5.94)} and semantics defined with the keyword *Boat* are shown in Fig. 3.

One of the biggest challenges when working with emotionally annotated databases IAPS and IADS is construction of scenarios in psychotherapy and psychological training. The scenarios have to be consistent and individually tailored for specific subjects and point in their therapy. Also, extraction of stimuli from databases has to be accurate, simple and fast. All these issues directly contribute to the quality of generated scenarios and success of their use.

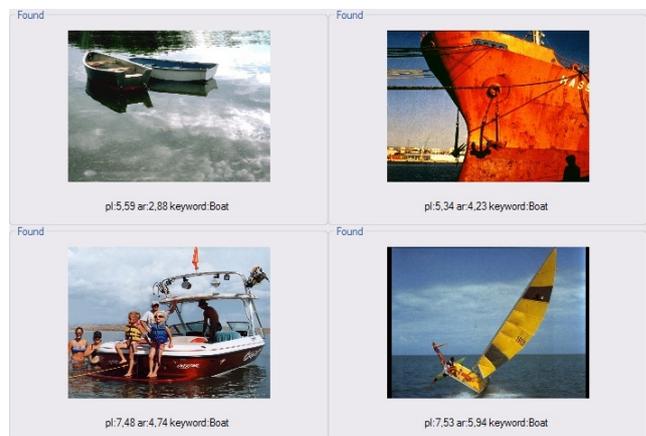

Fig. 3. Four visual emotionally annotated stimuli described with the same keyword *Boat*.

Content and purpose of scenarios in psychotherapy and psychological training can be defined only through harvesting individual emotional stimuli. Meaning of the scenario is an aggregation of semantics of all stimuli included in the scenario. In emotionally annotated databases semantics is exclusively encoded in stimuli keywords. However, distribution of keywords in IAPS and IADS databases with the respect to the stimuli is sparse and the structure is often inconsistent. Table 1 shows important statistical data on IAPS and IADS databases.

TABLE I
IAPS and IADS emotionally annotated databases statistics

| Feature | IAPS value | IADS value |
|---|---|---|
| Num. of stim. | 957 | 168 |
| Num. of keywords | 491 | 166 |
| Max. stim. per 1 keyword | 32 | 2 |
| Num. of keywords with 10+ stim. | 13 | 0 |
| Num. of keywords with 6-10 stim. | 10 | 0 |
| Num. of keywords with 4-5 stim. | 28 | 0 |
| Num. of keywords with 2-3 stim. | 100 | 2 |
| Num. of keywords with 1 stim. | 346 | 164 |
| Avg. num. of stim. per 1 keyword | 1.949 | 1.006 |
| Mode of stim. per 1 keyword | 1 | 1 |
| Median of stim. per 1 keyword | 1 | 1 |
| Std.dev. | 2.815 | 0.134 |

As can be seen in Table 1 the most well-known emotionally annotated databases (IAPS and IADS) have relatively many keywords compared to number of stimuli. This is especially noticeable in IADS statistics. In IAPS there is on average just 1.95 stimuli per single keyword. Median and mode measures of this ratio are both 1. Only 13 keywords define semantics for 10 or more stimuli, and 346 keywords define semantics for 1 stimulus. Distribution of stimuli per keywords in IADS is even more inept because almost every stimulus is described with its own unique keyword.

This data unequivocally shows the prevalence of describing stimuli semantics with unique keywords. Since keywords aren't semantically cross-linked in any machine-readable mode, semantics discovery is difficult. In their original form annotated databases demand a human expert that is well versed in their structure, emotional and semantic content.

### III. TAGS AND FOLKSONOMIES IN EMOTIONALLY ANNOTATED DATABASES

After examining emotionally annotated databases it becomes clear that the biggest problems in describing stimuli semantics with free-text keywords are:
1. Inconsistency
2. Ambiguity
3. Non-contiguousness

Inconsistency is demonstrated in various ways. For example simultaneous use in IAPS of singular and plural forms of the same semantic concept like with keywords *Woman/Women*, *Baby/Babies*, *Soldier/Soldiers* , or in concatenation of an adjective form and a noun, a noun and a verb, or even two nouns like in *AngryFace, GrievingFem, ManInPool, BoysReading, Girl&Dog*, etc. There is also simultaneous use of different terms in annotation of the same concept as in *Woman*, *Fem* and a suffix *–fem*, and lexical or appellation inconsistencies like *Cliffdiver* and *CliffDivers*, or *MenW/guns*. In addition to problems already mentioned there are incoherent pluralities and meanings like in *BikerCouple, Biking/train, NeuMan, NeutGirl, NeutralGirl, NeuWoman,* etc.

Ambiguity is a consequence of semantic inconsistency of the keywords. By analyzing IAPS and IADS tags it is not possible to make a unanimous decision whether some concepts are identical or different. Equally, it is impossible to numerically describe their mutual likeness. The tags do not provide enough information to completely describe and successfully discriminate stimuli semantics.

Non-contiguousness is also a consequence of the inconsistency. It is very difficult to define connections between IAPS tags based on their meaning, but as can be seen in the examples much if not all inconsistency could be avoided by storing nouns and verbs as unique concepts and assigning adverbs, adjectives and pluralities as their properties. Concepts could be conjoined with unary or binary operators (i.e. NOT, AND, OR, etc.) and formed into well-formed sentences. This ontological structure could be then used for automated reasoning by knowledge extraction with queries executed on the dataset.

Since emotionally annotated databases like IAPS and IADS are professionally recognized and widely used tools, it is important to also consider their keywords in the aspect of social indexing. Because these databases are used among different research groups it is plausible that at least some groups would like to modify tags or define their own tag sets. This would lead to new semantics and subsequently to the introduction of folksonomies [7]. Currently IAPS and IADS databases are not capable of supporting collaborative tagging: semantics of each stimulus is described with only one keyword, sets of stimuli and keywords are constant, and there is no mechanism or application that enables social tagging. Introduction of folksonomies in emotionally annotated databases would bring quantitatively and qualitatively better expressivity of stimuli semantics. This would improve the psychotherapy and psychological training and alleviate workload of the system's supervisor.

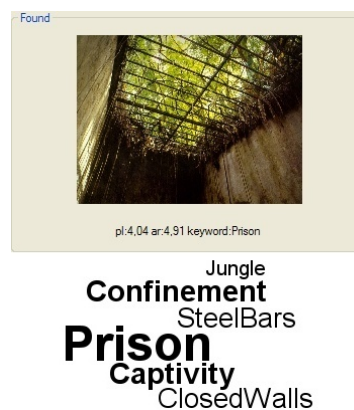

Fig. 4. IAPS visual stimulus *6000* and its tag cloud.

Another benefit of describing visual or auditory (i.e. multimedia) stimuli with multiple tags is a more accurate description of stimuli semantic content. An example is Fig.

4. Meaning of this IAPS stimulus is described with keyword *Prison*, however one can notice more aspects or meanings in the picture such as *Confinement*, *Captivity*, *SteelBars*, *ClosedWalls*, *Jungle*, etc. These keywords also describe semantics of the picture. Since they specify additional meaning they can be called *secondary semantic descriptors* or *secondary tags*. Primary description is the existing IAPS keyword, i.e. *Prison*, while all other keywords are auxiliary and should be specified in a social process by other IAPS users and organized in tag clusters. Importantly, in this process description value of secondary tags could be weighted and the tags sorted by their descriptive importance. Additional tags are important because IAPS keyword alone is not enough to relay all relevant information in the picture. Good example of this can be seen in Fig. 3. where the pictures are clearly different although they are tagged with the same keyword *Boat*. Furthermore, it should be possible to organize stimuli folksonomies into overlapping sets like semantic descriptors as tag clouds at image-hosting Web service Flickr [8]. In this way different stimuli can share the same subset of tags, and stimuli can be connected based on their common meaning.

To recapitulate: all described improvements can be achieved by expanding semantic description of a stimulus from one tag to a set of tags and organizing them into clusters and weighting as tag clouds. Finally, such tag sets should be arranged into folksonomies through an interactive social process.

## IV. CONSTRUCTING AN ONTOLOGY

While applying additional tags for description of multimedia emotional stimuli content and ordering them in tag clouds has helped in clarifying their meaning and increased their value in psychotherapy and psychological training, there is still no hierarchy or functional dependence between the tags. To overcome this problem it is necessary to transform tags into ontologies; establish their classes, properties, constraints and axioms, and also to create instances of ontology classes [9].

The first step in this process would be an application of taxonomy, i.e. classification of tags into a unified hierarchy with supertype-subtype (parent-child) mutual relationships. To this end we decided to use WordNet lexical database of English language [10]. Our decision was motivated by the size of WordNet (over 150,000 words) which guarantees that every keyword in emotionally annotated databases can be found in WordNet, and also by the quality of its organization and software implementation. Our motive was to pair every keyword with a noun, verb, adjective or adverb in WordNet, and use its knowledge structure (particularly *IS A* relationships and hypernym/hyponym hierarchy) to obtain a tree-like structure of IAPS and IADS keyword semantics. By definition *Y* is a hypernym of *X* if every *X* is a (kind of) *Y*, and by the same definition *X* a hyponym of *Y*. An example in Fig. 5 demonstrates complete WordNet hypernym and hyponym hierarchies for the keyword *Prison*

Any noun in the hypernym hierarchy of a keyword also describes the stimulus. In our example *correctional institution*, *penal institution*, *institution*, *establishment*, etc. describe the visual stimulus just along with its primary tag *prison*. However, it is very important to observe the semantic distance between the primary tag and individual nouns in the hypernym tree: the distance grows with the increase of number of nodes between two tags. For example, since all nouns in WordNet converge in synonym *entity*, it would be useless to differentiate noun tags in annotated databases with this symbol. In contract to hypernym hierarchies which provide taxonomy of terms, hyponym hierarchies can be used to provide variations and different versions of a keyword. They can be useful in broadening the keyword vocabulary which can better suite a wider range of users with diverse cultural backgrounds or language habits.

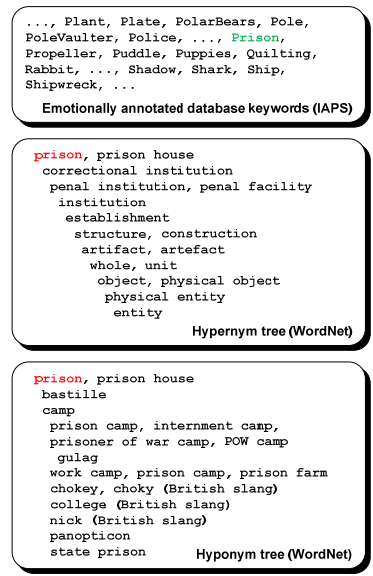

Fig. 5. Hypernym hierarchy and synonym matching for keyword *Prison*.

We can extract even more useful information from WordNet by coordinating terms with the primary tag. See Fig. 6. By definition *Y* is a coordinate term of *X* if *X* and *Y* share a hypernym. Thus, by term coordination it is possible to extract synonyms that are potentially similar to a specific keyword. They can be used as secondary semantic descriptors but one has to use them carefully since their informative value varies as it is strongly related to the content of a stimulus. In an overall performance terms coordination yields similar information as hyponym hierarchies.

The hypernym/hyponym relationships among the noun synonyms in WordNet can be interpreted as specialization relations between conceptual categories. However, such ontology cannot be directly transferred from WordNet and used in a custom application, but rather it has to be corrected because it contains inconsistencies such as the existence of common specializations for exclusive categories and redundancies in the hypernym hierarchy.

As can be seen in both Fig. 5 and Fig. 6 there are a number of potentially unwanted terms in the hierarchies which have to be removed before any use of WordNet as a source of relevant information. An example of this can be terms *training school*, *panopticon* or *reformatory* which may not suitable for describing semantics of the keyword *prison*, or *artifact*, *unit*, *physical entity*, *entity* that are largely or even completely irrelevant. Also, we had to dissect certain IAPS and IADS keywords which pair nouns, nouns and verbs, nouns and adjectives and verbs and adjectives (see examples in chapter III) into ontology class constraints or functions between classes.

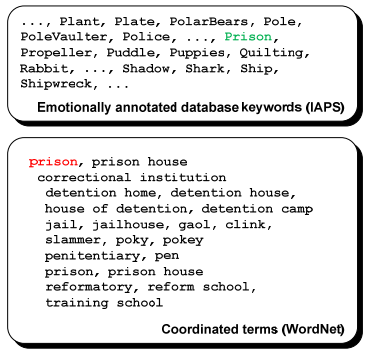

Fig. 6. Coordinated terms and synonym matching for keyword *Prison*.

Furthermore, in construction of our own ontology we noticed that taxonomy provided by WordNet is not expressive enough – especially for the purpose of PTSD diagnostics. Concepts represented in stimuli have multiple aspects to different subject who perceive them. Successful description of the meaning of stimuli and their mutual interaction is decisive for the processes of psychotherapy and psychological training. Thus, we had to describe and encode possible functional interactions between ontology concepts. In PTSD therapy of war traumatized patients which represents one of the primary target social groups in our research it is important to specify concepts like *person*, *family*, *soldier*, *civilian*, *man*, *woman*, *child* and bring them in connection with concrete and specific concepts like *army*, *terrorist*, *weapon*, *gun*, *sniper*, *shell*, *explosion*, *injury*, *death*, *torture*, *mutilation* and abstract concepts such as *joy*, *happiness*, *sorrow*, *fear*, *panic*, etc. In our continuing work with emotional stimuli ontologies we would like to describe knowledge such as: *Solider is a member of an army*, *Soldier can fire a weapon causing an injury or death*, *Soldiers sometimes terrorize people*, *Injury or death in most situation cause panic*, *In some cases injury causes mutilation, People can panic*, *If not in shock person is in a state of sorrow if is a member of his family is wounded or killed*, *If a person is wounded he usually feels fear*, etc. As can be seen these expressions are not explicit or crisp. This fuzzy knowledge must be adequately described and stored in the knowledge base. However, in our project it is not necessary to model belief, certainty or probability in facts and rules: StimGen application is not designed to help a psychotherapist in his decision-making process, but only in provoking therapeutically appropriate emotional responses from human subjects. Therefore, ontologies are used in a limited scope only to describe general meaning of annotated emotional stimuli, and not the entire expert knowledge about various psychological or mental states and conditions like PTSD. However, the described task of knowledge description is still very large and requires improvements and continuous additions to the developing ontology.

In our work we used Protégé tool and manually coded ontologies in OWL Web Ontology Language [11]. Although RDF (Resource Description Framework) would be sufficient for description of IAPS/IADS keyword concepts and WordNet generated taxonomy, we decided to use OWL rather than RDF because of OWL's greater expressivity. In the initial requirements analysis it was recognized that the ontology will be iteratively improved and it would be wrong to restrict the system from the start with the choice of inherently limited knowledge representation language. Also, the best option was to utilize the same contemporary language during the entire project. This facilitates exchange and reuse of the ontology. A snippet from an ontology based on aggregation of IAPS and IADS keywords which have been transformed into concepts and then mapped onto WordNet synonyms with addition of functional concepts is demonstrated in the figure below.

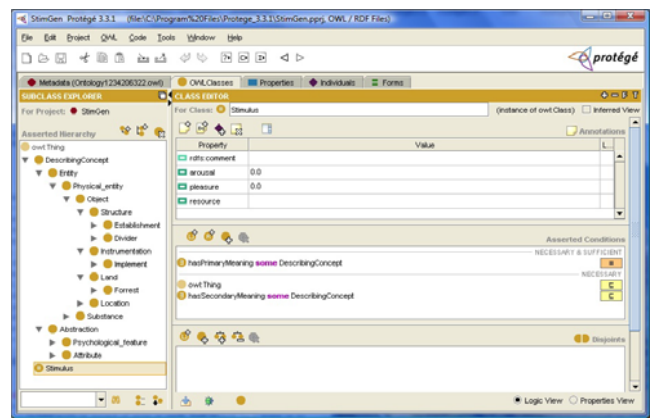

Fig. 7. Example of ontology for description of IAPS stimulus.

The ontology has two main classes: *Stimulus* and *DescribingConcept*. *Stimulus* has three annotation datatype properties: *arousal* (float), *pleasure* (float) and *resource* (string) that carry information about the emotional content according to the quantitative-dimensional psychological model and the stimulus resource file. Class *DescribingConcept* subsumes all concepts derived from IAPS/IADS keywords and WordNet lexical database which are relevant in describing the meaning of stimuli. These concepts are ordered by parent/child and other functional relations. Furthermore, there are two main pairs of object properties: *hasPrimaryMeaning/isPrimaryMeaningOf* and *hasSecondaryMeaning/isSecondaryMeaningOf*. Each stimulus is required to have some *hasPrimaryMeaning DescribingConcept*, and can have at least one *hasSecondaryMeaning DescribingConcept*. With this definition it is ensured that each stimulus is paired with at least one concept that describes its meaning.

## V. ATTACHING ADDITIONAL INFORMATION

In order to make the most of the process of introducing ontologies for knowledge description in emotional annotated stimuli, in addition to the steps already mentioned we also decided to use the Simple level of

Dublin Core metadata element set [12] for description of stimuli resource files. Image, sound or video files are a part of emotionally annotated databases, however their properties such as title, creator, subject, description, contributor, create date, type and format are unknown or must be derived implicitly. For example: IAPS stimulus 4000.jpeg has identification *4000*, its type derives from the stimulus' extension as *image/jpeg* and its description is hard-coded in the CSV manifesto file as *Artist*. As can be seen this resource information is sparse and it makes little contingencies for collaborative or cross-domain exchange.

Since Dublin Core metadata element set is a standard for cross-domain information resource description it can ensure simple and standardized set of conventions for description of resources in emotional annotated databases such as IAPS and IADS.

Fig. 8 demonstrates one proposal on how to encode information about one resource in RDF format; *dc:identifier* tag is used to store stimulus unique name and *dc:subject* its free-text keyword or keywords, *dc:type* describes the content of the resource by indicating emotionally annotated database to which the stimulus belongs, *dc:creator* refers to a person or body responsible for the content of the resource, *dc:contributor* is a person or entity responsible for making contributions to the content of the resource, *dc:date* is associated with an event in the lifecycle of the resource, and finally *dc:format* designates digital manifestation of the resource.

```
<?xml version="1.0"?>
<!DOCTYPE rdf:RDF PUBLIC "-//DUBLIN CORE//DCMES DTD
2002/07/31//EN"
    "http://dublincore.org/documents/2002/07/31/
dcmes-xml/dcmes-xml-dtd.dtd">
<rdf:RDF xmlns:rdf="http://www.w3.org/1999/02/22-rdf-
syntax-ns#" xmlns:dc="http://purl.org/dc/elements/
1.1/">
  <rdf:Description>
    <dc:identifier>4000</dc:identifier>
    <dc:creator>University of Florida</dc:creator>
    <dc:subject>Artist</dc:subject>
    <dc:contributor>Laboratory for interactive
simulation systems, Faculty of Electrical Engineering
and Computing</dc:contributor>
    <dc:date>2008-09-30</dc:date>
    <dc:type>IAPS</dc:type>
    <dc:format>Image</dc:format>
  </rdf:Description>
</rdf:RDF>
```

Fig. 8. Simple level of Dublin Core metadata elements in description of IAPS stimulus *4000*.

## VI. CONCLUSION

Current methods of content description in emotionally annotated databases have a low level of expressivity and accuracy. Indeed, there are many better paradigms for data storing and extraction which are used in other areas of computing such as relational databases or data warehouses. However, the best knowledge description procedures for the task of eliciting emotional response must be not only informatively rich but also self-explanatory and technically simple. Ontology is an excellent solution to this problem because by its definition [9] it represents a formal explicit specification of a shared conceptualization. It presents a shared understanding of knowledge structure and a common language between human experts and intelligent agents. Also, ontology facilitates its own reuse.

We started by augmenting predefined free-text keywords with folksonomies and broadening them to tag clouds, but we found ontology to be as yet the best tool for describing and harvesting information in emotionally annotated multimedia resources.

The future work will include knowledge description in video stimuli and improvement of the existing ontology through cooperation with domain experts, i.e. psychiatrists and psychologists. Our desire is to construct a common ontology that can be used for description of stimuli content in IAPS, IADS or any other *ad hoc* emotionally annotated database, regardless of its resource format – image, sound, video or even synthetic and VR environments. Such versatile StimGen can become a useful tool for the therapist for delivering multimedia stimuli to the patient during psychotherapy. We will also continue our work on integrating StimGen into the entire physiology-driven adaptive VR system for psychotherapy and psychological training.